\documentclass[10pt,twocolumn,letterpaper]{article}

\usepackage{iccv}
\usepackage{times}
\usepackage{epsfig}
\usepackage{amsmath}
\usepackage{amssymb}

\usepackage{graphicx}
\usepackage{booktabs}
\usepackage{bbding}
\usepackage{xcolor}

\usepackage{multirow}
\usepackage[noend]{algpseudocode}
\usepackage{algorithmicx,algorithm}
\usepackage{wrapfig,lipsum,booktabs}
\usepackage{algorithmicx,algorithm}

\newcommand{\z}{{\rm\bf z}}                   %
\newcommand{\w}{{\rm\bf w}}                   %
\newcommand{\txt}{{\rm\bf t}}                   %
\newcommand{\G}{{\rm\bf G}}                   %
\newcommand{\Z}{\mathcal{Z}}                  %
\newcommand{\W}{\mathcal{W}}                  %
\newcommand{\x}{{\rm\bf x}}                   %
\newcommand{\g}{{\rm\bf g}}                   %
\newcommand{\X}{\mathcal{X}}                  %
\newcommand{\Loss}{\mathcal{L}}               %
\newcommand{\blfootnote}[1]{
\begingroup
\renewcommand\thefootnote{}\footnote{#1}
\addtocounter{footnote}{-1}
\endgroup
}

\usepackage[pagebackref=true,breaklinks=true,letterpaper=true,colorlinks,bookmarks=false]{hyperref}

\iccvfinalcopy %

\def\iccvPaperID{9123} %

\ificcvfinal\fi

\begin{document}

\title{ManiCLIP: Multi-Attribute Face Manipulation from Text}

\author{Hao Wang$^1$ \quad 
Guosheng Lin$^{1\dagger}$ \quad 
Ana García del Molino$^2$  \quad
Anran Wang$^2$ \\
Jiashi Feng$^2$ \quad
Zhiqi Shen$^{1\dagger}$\\
$^1$Nanyang Technological University \quad $^2$ByteDance\\
}

\maketitle

\ificcvfinal\thispagestyle{empty}\fi

\blfootnote{$^\dagger$Corresponding authors}

\begin{abstract}
   In this paper we present a novel multi-attribute face manipulation method based on textual descriptions. Previous text-based image editing methods either require test-time optimization for each individual image or are restricted to single attribute editing. Extending these methods to multi-attribute face image editing scenarios will introduce undesired excessive attribute change, e.g., text-relevant attributes are overly manipulated and text-irrelevant attributes are also changed. In order to address these challenges and achieve natural editing over multiple face attributes, we propose a new decoupling training scheme where we use group sampling to get text segments from same attribute categories, instead of whole complex sentences. Further, to preserve other existing face attributes, we encourage the model to edit the latent code of each attribute separately via an entropy constraint. During the inference phase, our model is able to edit new face images without any test-time optimization, even from complex textual prompts. We show extensive experiments and analysis to demonstrate the efficacy of our method, which generates natural manipulated faces with minimal text-irrelevant attribute editing. Code and pre-trained model are available at \url{https://github.com/hwang1996/ManiCLIP}.
\end{abstract}

\section{Introduction}
Face editing \cite{xu2022transeditor,jiang2021talk,patashnik2021styleclip} aims to manipulate the given face images under certain instructions.
In this paper, we are interested in multi-attribute face editing, where one can change several different facial attributes in one shot, from the provided textual descriptions. 
Multi-attribute face editing is of great significance when users want to change their photo contents, including makeups, hair styles and face shapes. This gives people freedom to change the images and generate their desired images. However, relevant studies on multi-attribute editing are still rare.

\begin{figure}
\begin{center}
\includegraphics[width=0.48\textwidth]{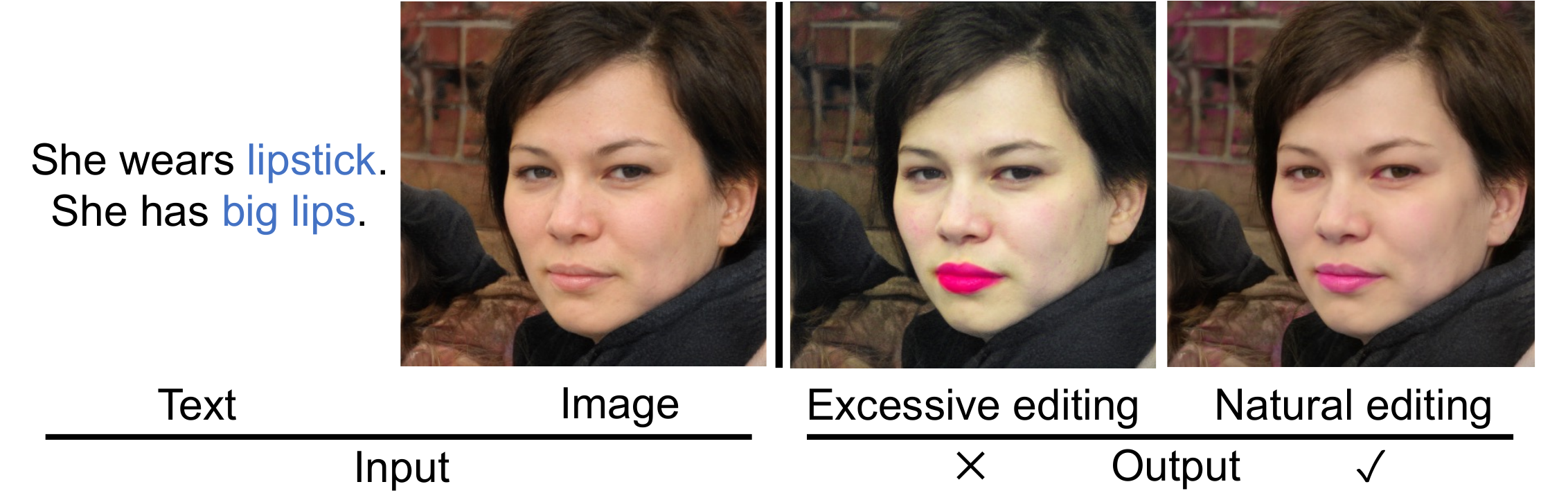}
\end{center}
\vspace{-10pt}
  \caption{\textbf{Comparison between excessive editing and natural editing.} The excessive and natural editing results are generated by our baseline model and our method respectively.}
\vspace{-10pt}
\label{fig:demo}
\end{figure}

With the development of StyleGAN2 \cite{karras2019style,Karras2019StyleGAN2} and CLIP \cite{radford2021learning} model, some recent works explore leveraging them for image content editing.
Specifically, StyleGAN2 has been demonstrated to learn disentangled latent codes \cite{karras2019style}, which are shown to have corresponding semantic meanings as their generated images \cite{xia2021tedigan,wang2022high}. CLIP model is pretrained on large-scale image-text datasets, which can measure the similarity between given images and text, by mapping them to the learned feature space. Therefore, to achieve automatic natural language based face editing, many works \cite{patashnik2021styleclip,xia2021tedigan,wei2022hairclip,ling2021editgan,xu2022transeditor} designed manipulation components to change the latent codes based on StyleGAN2 and CLIP model.

To be specific, TediGAN \cite{xia2021tedigan} uses CLIP loss \cite{radford2021learning} to supervise the text-image alignment of the edited images. This method suffers the problem of slow inference speed, as it requires individual alignment optimization over each image. StyleCLIP \cite{patashnik2021styleclip} proposes to train a mapper module to predict the proper offsets over the image latent codes to achieve desired editing.
However, they need to train various mappers for different text inputs. HairCLIP \cite{wei2022hairclip} follows the mapper design of StyleCLIP \cite{patashnik2021styleclip}, and only trains one mapper for different hair-related text inputs. But the original HairCLIP \cite{wei2022hairclip} is only applicable to single hair-related attribute editing, and the problem of multi-attribute editing remains unsolved.

\begin{figure*}
\begin{center}
\includegraphics[width=0.9\textwidth]{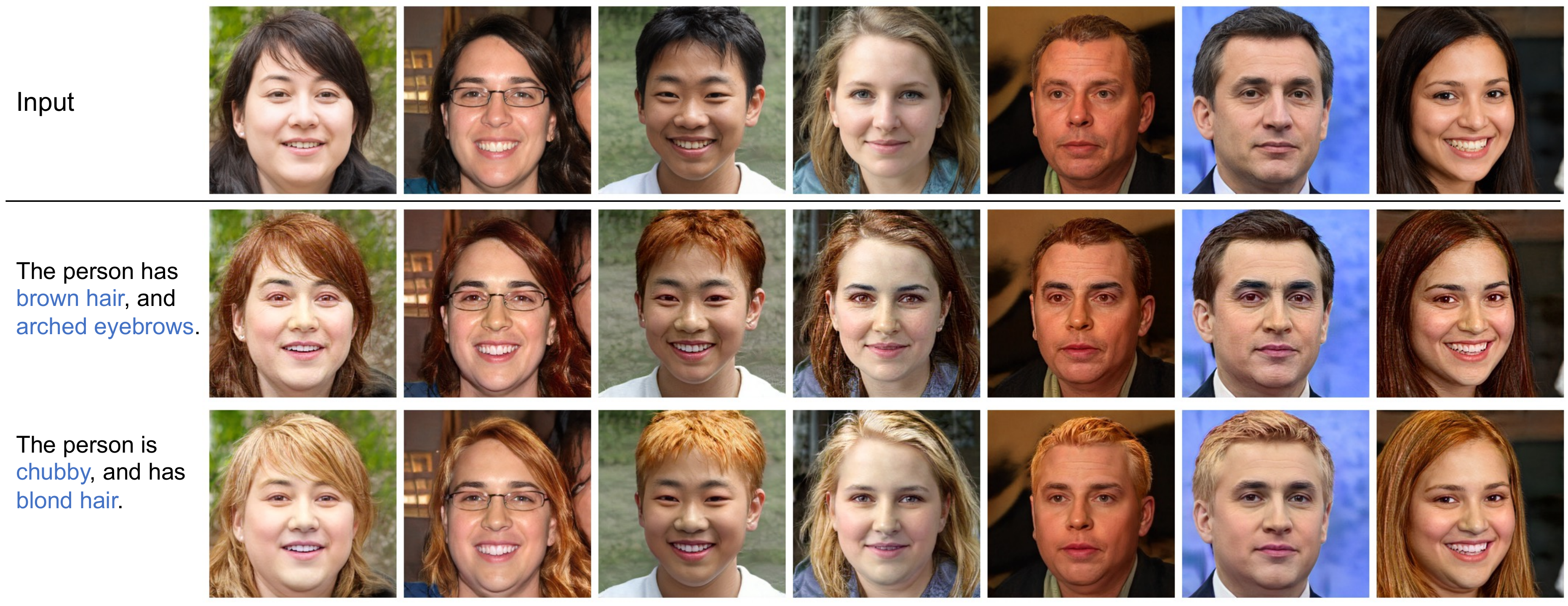}
\end{center}
\vspace{-10pt}
  \caption{\textbf{ManiCLIP:} our model is able to naturally edit multiple face attributes from natural language instructions. Top row shows the original images. Rows 2-3 show the edited face images under different text descriptions.}
\label{fig:teaser}
\end{figure*}

Though it is straightforward to extend previous methods to the general multi-attribute face manipulation task, we observe that the baseline model results in excessive editing, as shown in Figure \ref{fig:demo}. Excessive editing means: 1) the relevant attributes are overly amplified and unnatural results are produced, e.g. the lipstick has over bright color in Figure \ref{fig:demo}, and 2) the text-irrelevant attributes are also changed in the edited images, e.g. the skin color in Figure \ref{fig:demo}. 
To alleviate these issues, we introduce a new method, named ManiCLIP, that deploys a new decoupling training scheme and entropy constraint based loss design to generate natural edited images while minimizing the text-irrelevant attribute change.

Our proposed method is two-pronged.
First, 
we observe that the difficulty of attribute editing for the model is heterogeneous. For example, manipulation of \emph{lipstick} is easy, several epoch training would give overly edited results, while editing of other attributes may require more epochs. When we do mixed training, since ``hard" attributes are harder to synthesize, the model tends to keep optimizing the CLIP loss and, as a result, ``easy" attributes will become unnatural, for example the \emph{lipstick} in Figure \ref{fig:demo} presents excessive editing. 
Hence we propose the decoupling training scheme, where we use group sampling and  only edit one kind of attributes in each instance. It allows the model to fit each kind of attributes individually, which alleviates the issue of unnatural edited results.
Second, because of the disentanglement property \cite{Karras2019StyleGAN2} of StyleGAN2 latent space, only a small portion of the latent code dimensions affect certain attributes. However, StyleGAN2 latent codes have thousands of dimensions, which have much freedom during the editing process, hence we propose to use the entropy loss to control the number of non-zero dimensions. Since the uniform distribution of latent code values yields maximum entropy, minimizing the entropy loss can force the model to produce more values closer to zero, as visualized in Figure \ref{fig:entropy_demo}. During the training phase, we optimize the entropy loss and CLIP loss simultaneously, text-irrelevant attributes can be preserved and relevant attributes can be edited.

The comprehensive experimental results show the suitability of our proposed method avoiding test-time optimization. Based on textual descriptions containing multiple attributes, we generate natural manipulated images with minimal text-irrelevant editing by our proposed decoupling training scheme and entropy loss. We show the teaser in Figure \ref{fig:teaser}. 
Our contributions can be summarized as: 
\begin{itemize}
\vspace{-5pt}
   \item Adoption of a decoupling training scheme to enforce minimal overly editing for input text, which is beneficial to giving natural results.
   \vspace{-5pt}
   \item Application of the entropy loss onto the StyleGAN2 latent code offsets, which regularizes the offset freedom to avoid unnecessary change.
   \vspace{-5pt}
   \item We demonstrate that our proposed ManiCLIP outperforms several state-of-the-art methods for multi-attribute face editing task. %
\end{itemize}

\section{Related work}
\subsection{Face editing}
The challenge of face editing \cite{tan2020michigan,shen2020interpreting,saha2021loho,nitzan2020face,jo2019sc} is to change part of the attributes of the original face image, while preserving all other irrelevant attributes. Since the latent space $\W$ of StyleGAN2 is claimed to better reflect the disentangled semantics of the learned distribution \cite{Karras2019StyleGAN2,wu2021stylespace}, previous methods \cite{patashnik2021styleclip,xia2021tedigan,wei2022hairclip,jiang2021talk} take the architecture of StyleGAN2 \cite{Karras2019StyleGAN2} and do manipulation on its latent codes, such that the generated images can be edited and meet the desired outputs. 

Specifically, Talk-to-Edit \cite{jiang2021talk} introduces a dialog system to iteratively edit images, which adopts a pretrained attribute predictor to supervise the editing pipeline. Xu et al. \cite{xu2022transeditor} propose dual latent spaces based on the original StyleGAN2 architecture, and claim that their proposed cross-space interaction allows cooperated complex editing. However, this method is dependent on InterFaceGAN \cite{shen2020interpreting}, which requires individual operation over each sample. 

\begin{figure*}
\begin{center}
\includegraphics[width=0.8\textwidth]{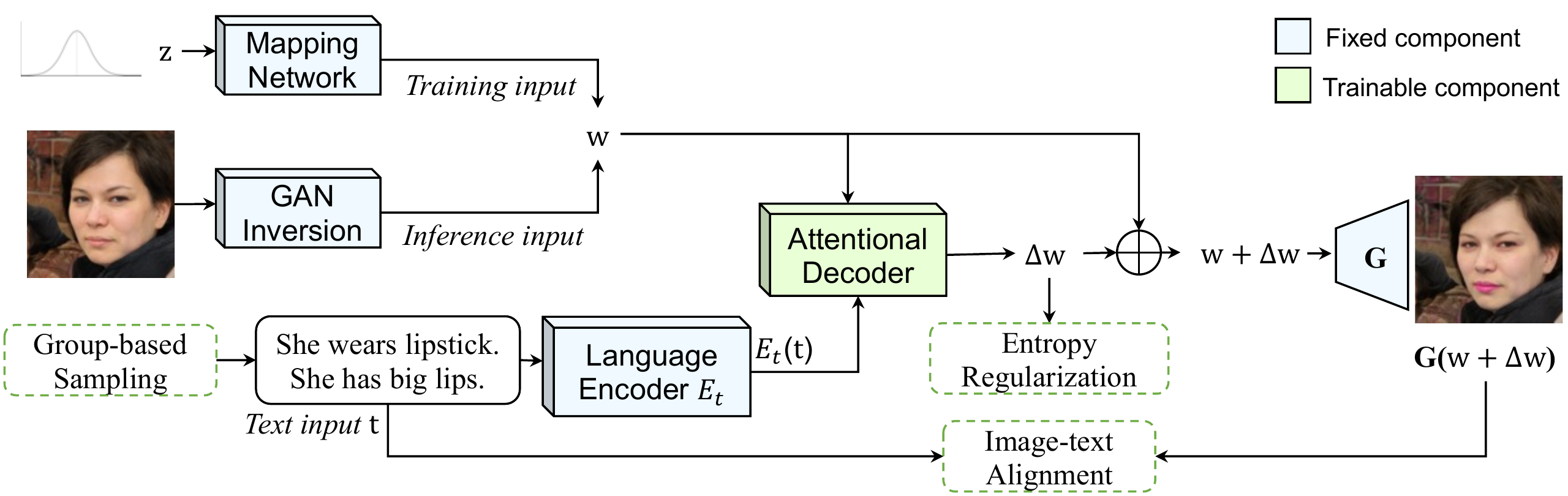}
\end{center}
\vspace{-10pt}
  \caption{\textbf{ManiCLIP: our proposed multi-attribute face editing model.} Our model consists of several modules: a StyleGAN2 and mapping network pretrained on FFHQ dataset \cite{karras2019style}, a CLIP \cite{radford2021learning} language encoder $E_t$, a transformer-based attentional decoder, and a GAN inversion model \cite{tov2021designing}. This architecture enables us to generate natural manipulated images with minimal text-irrelevant editing.}
\label{fig:framework}
\end{figure*}

TediGAN \cite{xia2021tedigan}, StyleCLIP \cite{patashnik2021styleclip} and HairCLIP \cite{wei2022hairclip} all adopt CLIP \cite{radford2021learning} model to achieve text-image semantic alignment. StyleCLIP \cite{patashnik2021styleclip} proposes both an optimization-based and mapper-based method to produce edited images with the pretrained CLIP model \cite{radford2021learning}, where the former one requires specific optimization over latent codes and the latter requires training different models for different input text.
TediGAN \cite{xia2021tedigan} follows an optimization-based approach, but is not restricted to text-guided manipulation: sketch or semantic masks can also be used as conditions to control the output. Technically, they first get the inverse latent codes of real images, and train an additional encoder to align the sketch or mask input with the corresponding StyleGAN2 latent codes. The produced latent codes from the trained encoder then requires further optimization, which is slow at inference time. HairCLIP \cite{wei2022hairclip} adopts a mapper-based method, which focuses on manipulating the hair styles, but it can only change one attribute during each editing time.

To summarize, adopting CLIP \cite{radford2021learning} as the main supervision signal for the text-guided face editing task is useful. In this paper, we aim to change multiple face attributes simultaneously, where ones do not need to further optimize the edited results. We follow previous practise to use CLIP for supervision, and additionally introduce decoupling training strategy and entropy loss to achieve minimal excessive editing for text segments.

\subsection{Text-based image generation and manipulation}
Both text-based image generation \cite{zhou2022towards,xu2018attngan,zhang2021cross,zhu2019dm,ding2021cogview,ramesh2021zero} and manipulation \cite{li2020manigan,patashnik2021styleclip,tov2021designing,collins2020editing,chen2018language,li2019controllable} tasks aim to generate images from the text conditions. For image manipulation, it also requires high similarity between manipulated and original images. To this end, many research works \cite{li2020manigan,xu2018attngan,zhu2019dm,cheng2020rifegan} utilize the conditional GAN structure, where the input text representations are used as the conditional information of GAN, and add various regularization to enhance the generation quality. Since these methods \cite{li2020manigan,xu2018attngan,cheng2020rifegan} train the whole framework in an end-to-end manner and use a discriminator only to improve the image quality, the generated images may not exactly match with the given text and the generation diversity is limited \cite{wang2021cycle}.

With the emerging applications of StyleGAN2 \cite{Karras2019StyleGAN2}, recent works \cite{wang2021cycle,patashnik2021styleclip} adopt StyleGAN2-based method. For example, CI-GAN \cite{wang2021cycle} follows TediGAN \cite{xia2021tedigan} and proposes to use cycle consistency for GAN inversion. They demonstrate that unconditional training increases the diversity of GAN generation, and that doing manipulation on StyleGAN2 latent space can ensure the generation quality. Therefore, we also use a pretrained StyleGAN2 as our model backbone.

In DALL-E \cite{ramesh2021zero,ramesh2022hierarchical}, a transformer is used for text-to-image generation. Specifically, DALL-E-2 \cite{ramesh2022hierarchical} feeds the CLIP text embeddings to the model and produces the image embeddings first, then uses the diffusion decoder to generate the final images. However, they use about 650M images for training. In contrast, we only use a pretrained StyleGAN2 model and do not need any images during the training phase, since we achieve the multi-attribute face editing by randomly sampling StyleGAN2 latent codes.

\section{Method}

\subsection{Preliminary}
\noindent\textbf{StyleGAN2.} We denote StyleGAN2 \cite{Karras2019StyleGAN2} model as $\G: \Z\rightarrow\W\rightarrow\X$, where $\Z$, $\W$ and $\X$ represents noise space, learned latent space, and generated image space respectively. StyleGAN2 uses Multi-Layer Perceptron (MLP) to map the initial $\Z$ noise space to the $\W$ latent space. $\G$ generates images based on latent codes $\w \in \W$. StyleGAN2's learned $\W$ space presents a disentangled nature, where different attributes are controlled via different dimensions of $\w$ \cite{karras2019style,Karras2019StyleGAN2}.

\noindent\textbf{CLIP.} CLIP \cite{radford2021learning} model is trained with large-scale image-text datasets, containing 400 million pairs from the Internet. Both pretrained vision and language encoders are available. We use the features extracted using them to measure the cosine similarity between images and text.

\subsection{Overview}

A natural approach to multi-attribute face editing without test-time optimization would be to adapt StyleCLIP's \cite{patashnik2021styleclip} mapper design to multi-attribute text datasets. However, the adapted model may present excessively edited face images, as shown in Figure \ref{fig:demo}. In this paper, we aim to solve the excessive editing problem.

The pipeline of our proposed text-based multi-attribute face editing framework is presented in Figure \ref{fig:framework}. We train the attentional decoder only and fix all other pretrained components. Specifically, StyleGAN2 $\G$ and its mapping network are pretrained on FFHQ dataset \cite{karras2019style}, and $E_t$ is the CLIP \cite{radford2021learning} pretrained language encoder. During the training phase, we sample random noise $\z \in \Z$ and feed it into the StyleGAN2 mapping network to obtain the latent codes $\w \in \W$. To obtain the training text segments $\txt$, the decoupling training scheme is adopted, where we use group sampling. The text embeddings $E_t(\txt)$ are extracted with the CLIP language encoder $E_t$. Then $E_t(\txt)$ and $\w$ are fed into the transformer-based attentional decoder to generate the offset $\Delta\w$. 
The edited images $\G(\w + \Delta\w)$ are produced through a pretrained StyleGAN2 model $\G$, which is supervised by the image-text alignment module. We also apply entropy regularization onto the learned $\Delta\w$.

During the inference phase, we use a GAN inversion encoder \cite{tov2021designing} to get $\w$ from real images, and use complex sentences with multiple attributed as input text. $\Delta\w$ is directly used to generate the edited images, not requiring additional optimization.

\begin{figure}
\begin{center}
\includegraphics[width=0.32\textwidth]{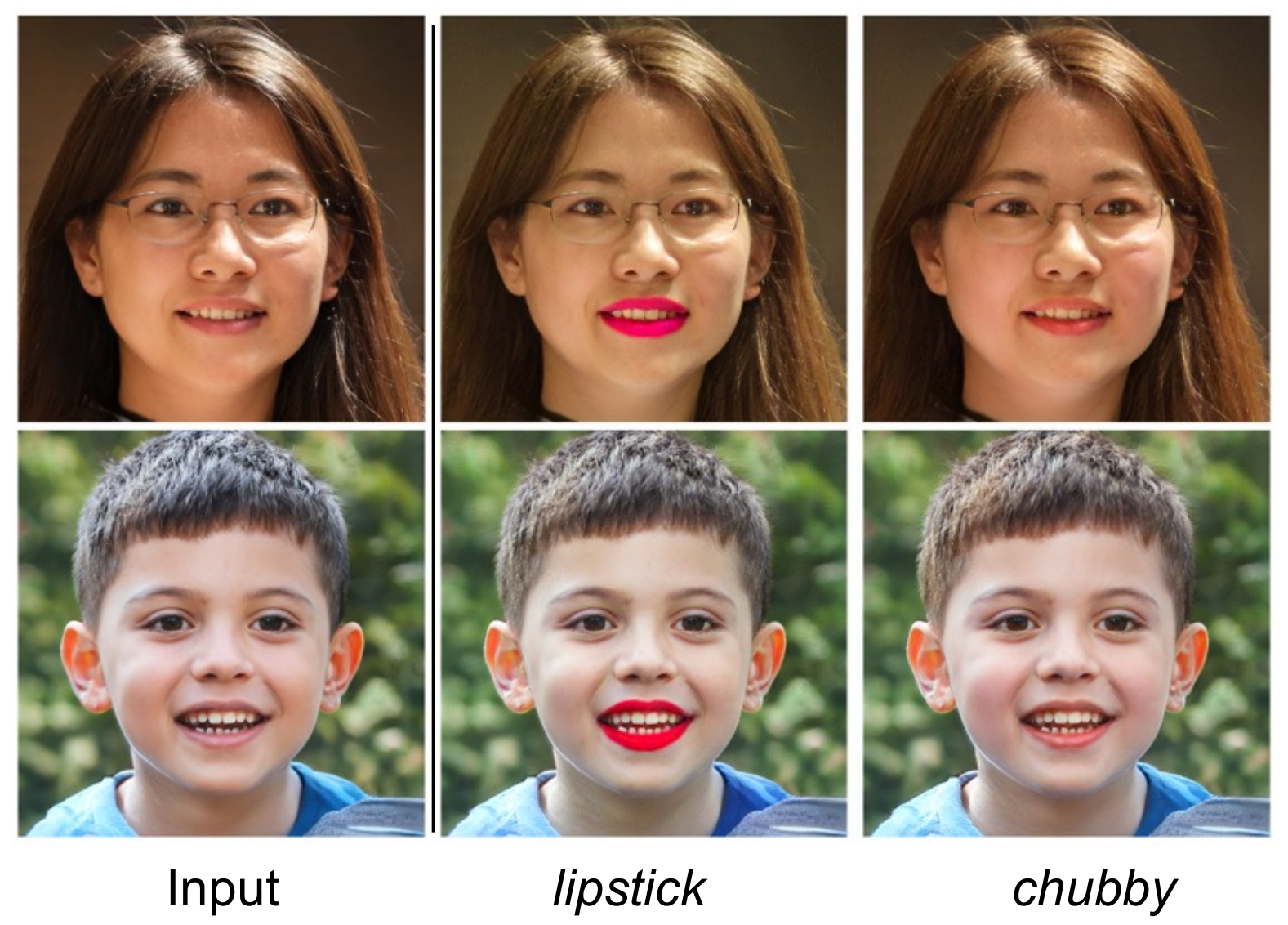}
\end{center}
\vspace{-10pt}
  \caption{We train the baseline model with 10 epochs, and show the edited results of \emph{lipstick} and \emph{chubby} respectively. The \emph{chubby} editing is harder than \emph{lipstick} editing, which is overly modified.}
\vspace{-10pt}
\label{fig:speed}
\end{figure}

\subsection{Image-text alignment}

We first study using a reasonable baseline that considers identity and background preservation.
To supervise the semantic consistency between the given text $\txt$ and $\G(\w+\Delta\w)$, we adopt CLIP loss $\Loss_{clip}$ , which can be denoted as
\begin{equation}
    \Loss_{clip} = 1 - cos(E_i(\G(\w+\Delta\w)), E_t(\txt)),
\end{equation}
where $E_i$ and $E_t$ represent the CLIP image and text encoder respectively, and $cos$ indicates the cosine similarity between two items.

To preserve face identity and background information, we use face ID loss \cite{deng2019arcface} $\Loss_{id}$, background loss $\Loss_{bg}$ and image L2 loss $\Loss_{l2}^i$.
$\Loss_{id}$ is utilized to ensure that the face identity remains unchanged, which is also used in previous face editing works \cite{patashnik2021styleclip,wei2022hairclip}. It is derived from the pretrained ArcFace \cite{deng2019arcface} $R$, and the loss can be depicted as
\begin{equation}
    \Loss_{id} = 1 - cos(R(\G(\w+\Delta\w)), R(\G(\w))).
\end{equation}
Using $\Loss_{bg}$ aims to keep the image background unchanged. To this end, we use a face segmentation model \cite{zhao2017pyramid} to obtain the background area of face images, then we apply the L2 loss on the background area. $\Loss_{bg}$ is denoted as
\begin{equation}
\begin{aligned}
    \Loss_{bg} = ||(\G(\w+\Delta\w)-\G(\w))*(M_{\Delta\w}^{bg} \cap M_{\w}^{bg})||_2,
\end{aligned}
\end{equation}
where $M_{\Delta\w}^{bg}$ and $M_{\w}^{bg}$ represent the background masks of edited and original images respectively, and $\cap$ denotes the union of $M_{\Delta\w}^{bg}$ and $M_{\w}^{bg}$.
$\Loss_{l2}^i$ is adopted to prevent unnecessary change of text-irrelevant attributes on face areas, and it is defined as
\begin{equation}
\begin{aligned}
    \Loss_{l2}^i = ||(\G(\w+\Delta\w)-\G(\w))*(M_{\Delta\w}^{face} \cap M_{\w}^{face})||_2.
\end{aligned}
\end{equation}
where $M_{\Delta\w}^{face}$ and $M_{\w}^{face}$ represent the face masks of edited and original images. Finally, to keep a general similarity between the manipulated and original images, previous methods \cite{patashnik2021styleclip,xia2021tedigan,wei2022hairclip} utilize the latent code norm loss $\Loss_{l2}^w$ to avoid $\Delta\w$ being too large, \ie 
\begin{equation}
    \Loss_{l2}^w = ||\Delta\w||_2.
\end{equation}

However, we observe that the baseline model trained with the aforementioned losses produces excessively edited images, as shown in Figure \ref{fig:demo}.
To resolve the excessive editing issue, we introduce a decoupling training scheme and apply an entropy constraint on the learned $\Delta\w$.

\subsection{Decoupling training scheme for natural editing}
The decoupling training scheme is designed based on our empirical observations on the model convergence speed, as shown in Figure \ref{fig:speed}.
Specifically, we define the ``easy" attributes as more local ones that take less training iterations to give visible editing results, and ``hard" attributes as more global ones that take more iterations for manipulation.
If we use sentences containing mixed ``easy" (e.g., \emph{lipstick}) and ``hard" (e.g., \emph{chubby})  attributes, 
since ``hard" attributes are harder to adjust, ``easy" attributes would be overly modified for CLIP loss optimization. 
To alleviate the unnatural editing issue, we propose to do decoupling training where we only try to edit one kind of attributes in each instance, instead of using the full input sentences containing mixed attributes. It allows the model to fit each kind of attributes individually, and generate more natural edited images.

Our proposed decoupling training scheme aims to put attributes with similar learning speed in single sentences. To this end, we empirically categorize the given 40 CelebA face attributes into 5 groups, i.e. \emph{hair}, \emph{eye}, \emph{mouth}, \emph{fashion} and \emph{others}. During the training phase, in each iteration, we randomly sample one category and pick random attributes under the sampled category. We then combine the sampled random attributes as a training text segment. Here the sampling attribute number $N_a$ is a hyper-parameter. 
We refer to the aforementioned setting as \textit{group sampling}. We also conduct random sampling for comparison purpose, where the attributes are sampled from all given 40 attributes, instead of specific attribute group.
We present the full procedure in Algorithm \ref{algo:1}.

By this way, we force the model to focus on a single attribute category instead of mixed categories during the training of each sample, so that the risk of ``easy'' attributes being overly edited is reduced. We achieve better manipulation results in terms of producing natural results with this decoupling training scheme.

\begin{algorithm}[t]
\caption{Decoupling training scheme} 
\label{algo:1}
\hspace*{0.02in} {\bf Input:}
All face attributes $\{\x_i\}_{i=1}^{40}$, attribute groups $\{\g_i\}_{i=1}^5$ and sampling attribute number $N_a$;
\begin{algorithmic}[1]
\State Initialize sampling strategy $\theta \in \{\text{random}, \text{group}\}$;
\If{$\theta$ is random}
    \State Sample $\{\x_i\}_{i=1}^{N_a}$ from $\{\x_i\}_{i=1}^{40}$; 
\EndIf
\If{$\theta$ is group}
    \State Sample one attribute group $\g_s$ from $\{\g_i\}_{i=1}^5$;
    \State Sample $\{\x_i\}_{i=1}^{N_a}$ from $\g_s$;
\EndIf
\State \Return concatenated sentences from $\{\x_i\}_{i=1}^{N_a}$.
\end{algorithmic}
\end{algorithm}

\subsection{Entropy constraint to avoid irrelevant change}
\begin{figure}
\begin{center}
\includegraphics[width=0.35\textwidth]{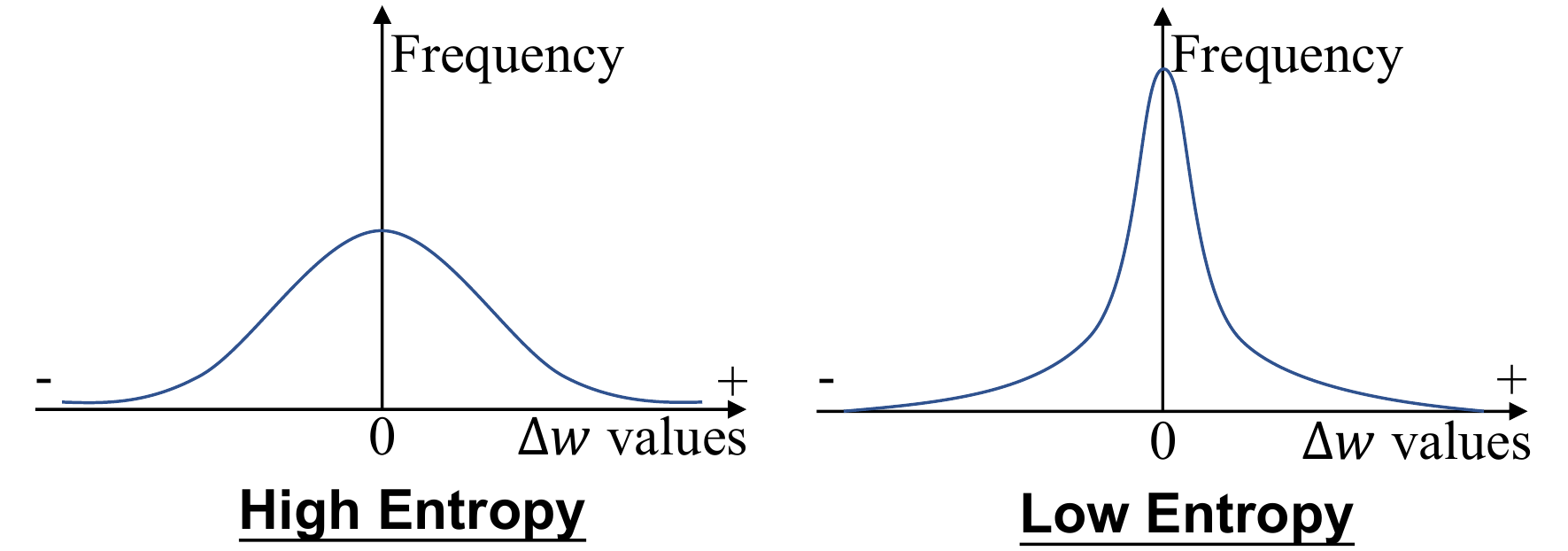}
\end{center}
\vspace{-10pt}
  \caption{\textbf{Demonstration of high and low entropy of $\Delta\w$ distribution.} The horizontal axis represents the value range of $\Delta\w$, and the vertical axis denotes the corresponding number of each $\Delta\w$ value. In the low entropy of $\Delta\w$, there are more zero or small values, which prevent the editing on text-irrelevant attributes.}
\vspace{-10pt}
\label{fig:entropy_demo}
\end{figure}

We assume $\w \in \W$ is disentangled \cite{Karras2019StyleGAN2}. Hence only part of the latent code $\w$ offset dimensions should have non-zero values, while all other dimensions should have small or zero values. Although the latent code norm loss constrains the offsets to some extent, it only gives general suppression on the offsets values, and using too much weight on latent code norm loss makes the CLIP loss optimization difficult. Therefore, we propose to apply entropy loss $\Loss_{en}$ on offset $\Delta\w$, which can be optimized simultaneously with CLIP loss. Minimizing $\Loss_{en}$ can not only force the model to give more zero values, but also maintain non-zero values, as shown in Figure \ref{fig:entropy_demo}. By doing this, text-irrelevant attributes can be preserved and relevant attributes can be edited.

We define the normalized $\Delta\w$ by $p(\Delta\w)$ as follows:
\begin{equation}
    p(\Delta\w) = \frac{||\Delta\w||}{\rm{max}(||\Delta\w||)},
\end{equation}
where $||\Delta\w||$ denotes the absolute values of $\Delta\w$, since both positive and negative values change the semantics of $\w$. It is notable that we do not use the softmax normalization to get $p(\Delta\w)$, as we observe softmax normalization changes the original distribution of $\Delta\w$ and yields inferior performance than our adopted min-max normalization method. 

The entropy loss $\Loss_{en}$ is defined by the Shannon Entropy formulation \cite{shannon1948mathematical}, and is denoted as 
\begin{equation}
    \Loss_{en} = -\sum_{i=1}^Np(\Delta\w)\rm{log}(\mathit{p}(\Delta\w)).
\end{equation}
where $N$ is the dimension number of $\Delta\w$. With the help of $\Loss_{en}$, we suppress text-irrelevant $\Delta\w$ dimension values, without affecting the relevant ones. This results in minimization on text-irrelevant attribute editing of the generated images.

\subsection{Training objective}
To summarize, our overall training objective is  
\begin{equation}
\begin{aligned}
    \Loss = &\Loss_{clip} + \lambda_{id}\Loss_{id} + \lambda_{bg}\Loss_{bg} + \lambda_{l2}^i\Loss_{l2}^i   \; \\
    &+\lambda_{l2}^w\Loss_{l2}^w + \lambda_{en}\Loss_{en},
\end{aligned}
\end{equation}
where $\lambda_{id}$, $\lambda_{bg}$, $\lambda_{l2}^i$, $\lambda_{l2}^w$ and $\lambda_{en}$ are set as 0.2, 1, 0.02, 0.1 and 0.2 respectively.

\section{Experiments} \label{sec:exp}
\subsection{Setup}
\noindent\textbf{Dataset.}
It is notable that our proposed method does not require any real images or text for training. We use the original 40-category face attributes from CelebA-HQ \cite{CelebAMask-HQ} dataset to construct the random training text.
We use text data from the Multi-Modal-CelebA-HQ dataset \cite{CelebAMask-HQ,xia2021tedigan} for testing purposes, which is generated based on the original annotations \cite{CelebAMask-HQ}. This dataset contains 30,000 image and text pairs, where each text file has 10 captions to describe face images. We use the first sentence of the last 5,000 text files for evaluation. 
During the testing phase, we use 5,000 fixed latent codes and sentences to output the edited images. 

\begin{figure*}
\begin{center}
\includegraphics[width=0.98\textwidth]{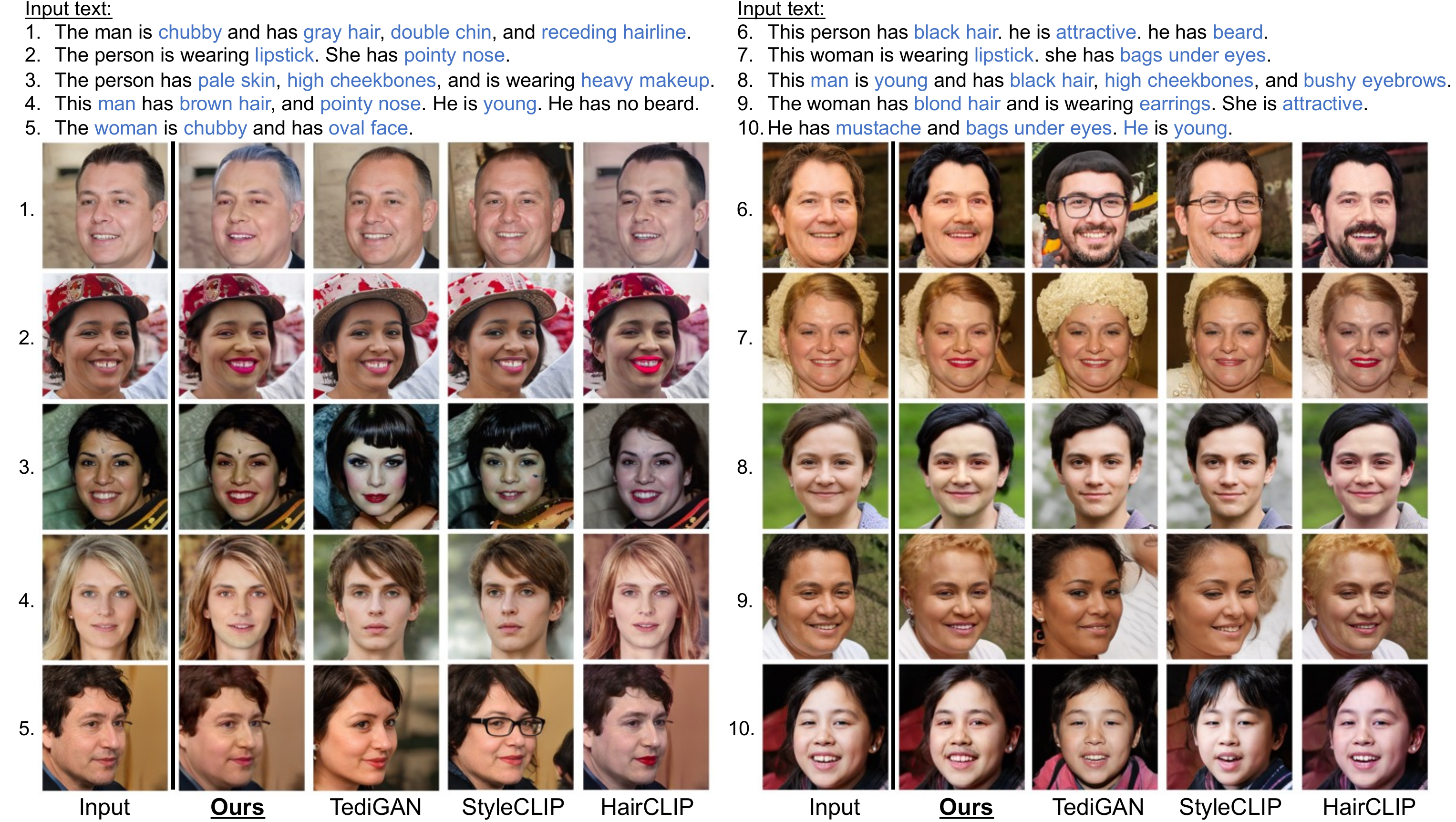}
\end{center}
\vspace{-10pt}
  \caption{\textbf{Comparison against existing methods.} The input text is shown in the top and the input face images are presented in the left columns. It can be seen that previous methods produce excessive editing on the manipulated faces, such as changed identities, different head poses, unnatural lipstick, etc. In contrast, our method generates natural edited images, with minimal text-irrelevant change.}
\vspace{-5pt}
\label{fig:against_others}
\end{figure*}

\noindent\textbf{Evaluation metrics.}
We use Fr\'echet Inception Distance (FID) \cite{heusel2017gans} to evaluate the edited image quality. We also use identity similarity (ID) to evaluate if the edited images keep the original identity information. Moreover, we train an individual face attribute model to predict attribute labels for face images. Based on the predicted labels, we calculate accuracy (Acc) to quantitatively evaluate if the model conducts excessive editing. To be specific,
we use the accuracy of edited image labels over target labels as Acc.

\noindent\textbf{Implementation details.} We use the StyleGAN2  model\footnote{https://github.com/rosinality/StyleGAN2-pytorch} \cite{Karras2019StyleGAN2} pretrained on FFHQ dataset \cite{karras2019style} as the image generator. The $\Delta\w$ decoder is a 6-layer transformer, where the head number is set as 8. We set the batch size and learning rate as 16 and 0.0001 respectively, and we use Adam \cite{kingma2014adam} optimizer. The training epoch number is 30. We use a single V100 for training.

\begin{table}
 \caption{\textbf{Quantitative comparison against existing works.} We use FID, ID similarity (ID) and attribute prediction accuracy (Acc) to do evaluations. ($^*$HairCLIP is our reproduced version, and hair-related loss is removed.)}
 \label{tab:others}
 \centering
 \resizebox{0.32\textwidth}{!}{
 \begin{tabular}{l|cccc}\toprule
\textbf{Models} &\textbf{FID $\downarrow$} & \textbf{ID $\uparrow$} & \textbf{Acc $\uparrow$}\\\midrule
TediGAN \cite{wei2022hairclip} &45.62 &45.46 &81.23 \\
StyleCLIP \cite{patashnik2021styleclip} &44.41 &44.77 &80.91 \\
HairCLIP$^*$ \cite{wei2022hairclip} &41.56 &62.41 &82.50 \\
ManiCLIP (ours) & \textbf{31.57} & \textbf{76.55} & \textbf{85.33} \\
\bottomrule
\end{tabular}
}
\vspace{-15pt}
\end{table}

\subsection{Comparison against other methods}
We show the quantitative results for our proposed method and various existing methods in Table \ref{tab:others}, where we evaluate them by FID, identity similarity (ID) and attribute prediction accuracy (Acc). It is notable that both TediGAN \cite{xia2021tedigan} and StyleCLIP \cite{patashnik2021styleclip} require individual test-time optimization over each instance, while HairCLIP \cite{wei2022hairclip} and ours do not require. Compared with previous works, our method has big improvements on the FID and ID similarity metrics, which indicates the natural editing of our proposed ManiCLIP. Acc measures the correctness of our edited results, demonstrating the efficacy of ManiCLIP on preventing text-irrelevant change. We observe ManiCLIP consistently outperforms previous methods across all the metrics.

We also present qualitative results in Figure \ref{fig:against_others}, where the input text is shown in the top and the input face images are shown in the left columns. To be specific, we can see that generally all edited images match the given text semantics. However, when we compare the manipulated images with the original ones, TediGAN and StyleCLIP change too much the face identities. For example, in the fourth instance, where the text indicates the gender is male, though TediGAN and StyleCLIP successfully produce images with a man, the generated images are dissimilar with the original image. In contrast, our edited image not only matches with the given text, but also has higher similarity with the original one, including the pose, hairstyle, expressions, etc. Comparing with HairCLIP (our reproduced version without hair-related losses), our method alleviates the issue of excessive editing. In the second example, HairCLIP's manipulated image has unnatural lipstick, and the skin colour is paler than the original, whereas our edited images are natural, and also have alignment with both original images and given text. The presented examples in Figure \ref{fig:against_others} demonstrate the importance of avoiding excessive editing in the face multi-attribute manipulation task, and our proposed decoupling training scheme and entropy loss alleviates excessive editing effectively.

\begin{table}
 \caption{\textbf{Evaluation of our proposed decoupling training scheme and entropy loss.} We use FID, ID similarity (ID) and attribute prediction accuracy (Acc) to do the evaluations.}
 \label{tab:ablation}
 \centering
 \resizebox{0.38\textwidth}{!}{
\begin{tabular}{cc|cccc}
\toprule
\textbf{Decouping} & \textbf{Entropy loss} & \textbf{FID $\downarrow$} & \textbf{ID $\uparrow$} & \textbf{Acc $\uparrow$} \\
\midrule
$\times$ & $\times$ &40.70 &69.88 &83.60 \\
$\times$ &\checkmark &38.62 &73.24 &84.73 \\
\checkmark & $\times$ &34.60 &74.41 &84.38 \\
\checkmark &\checkmark &\textbf{31.57} & \textbf{76.55} & \textbf{85.33} \\
\bottomrule
\end{tabular}
}
\end{table}

\begin{table}
 \caption{\textbf{Ablation study regarding the sampled attribute numbers $N_a$.} Random sampling denotes we sample $N_a$ attributes from all attributes, while Group sampling denotes we sample $N_a$ attributes out of same attribute groups. We use FID for evaluation.}
 \label{tab:decoupling}
 \centering
 \resizebox{0.45\textwidth}{!}{
 \begin{tabular}{l|ccccc}\toprule
\textbf{Attribute number $N_a$}  & \textbf{1} & \textbf{2} & \textbf{3} & \textbf{4} & \textbf{5}\\\midrule
\textbf{\textbf{Random sampling}} & 35.50 & 36.25 & 35.94 & 34.80 & 37.80 \\\midrule
\textbf{\textbf{Group sampling}} & 34.01 & 32.95 & \textbf{31.57} & 35.99 & 36.26 \\
\bottomrule
\end{tabular}
}
\end{table}

\begin{figure}
\begin{center}
\includegraphics[width=0.48\textwidth]{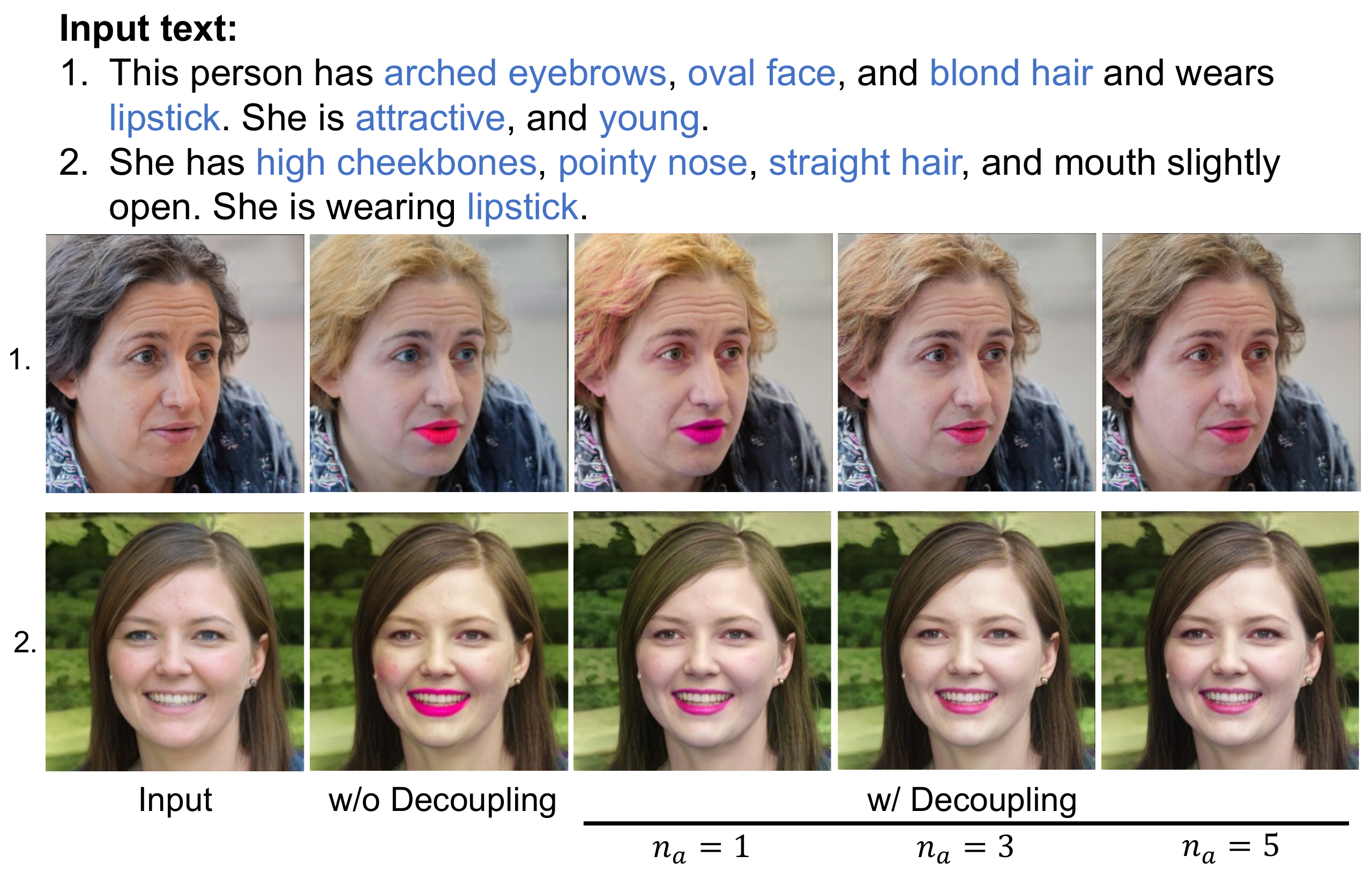}
\end{center}
\vspace{-15pt}
  \caption{\textbf{Effects of different attribute numbers $N_a$ sampled for each text segment in the decoupling training.} We use group sampling in the decoupling training scheme. In the decoupling setting, we show results of $N_a=1, 3, 5$ respectively.}
\vspace{-10pt}
\label{fig:decoupling_num}
\end{figure}

\subsection{Ablation study}\label{sec:ablation}
To demonstrate the usefulness of our proposed decoupling training scheme and entropy loss, we show the quantitative experiments in Table \ref{tab:ablation}. Specifically, only adopting decoupling training scheme already improves much on FID, showing it helps on giving more natural edited results. Using only the entropy loss brings more improvement on Acc, indicating it can preserve text-irrelevant attributes. Combining them together boosts the performance further.

\noindent\textbf{Analysis of decoupling training scheme.} 
To see the effect of the attribute number $N_a$ in the decoupling training, we conduct ablation study regarding different $N_a$. The quantitative results are presented in Table \ref{tab:decoupling}. We also give qualitative demonstrations in Figure \ref{fig:decoupling_num}.  
Using decoupling training scheme aims to generate natural editing results. 
Specifically, 
using $N_a=3$ with group sampling yields the best quantitative results.
In Figure \ref{fig:decoupling_num}, we observe the model without decoupling training scheme gives unnatural \emph{makeup} in both cases. We take the first instance of Figure \ref{fig:decoupling_num} as an example: when $N_a=1$, the edited hair color has even some pink shade, which is undesirable; when $N_a=5$, the edited hair color remains dark. Setting $N_a=3$ not only meets the text information, but also maintains minimal excessive editing. Hence we refer this setting as optimal.

\begin{figure}
\begin{center}
\includegraphics[width=0.42\textwidth]{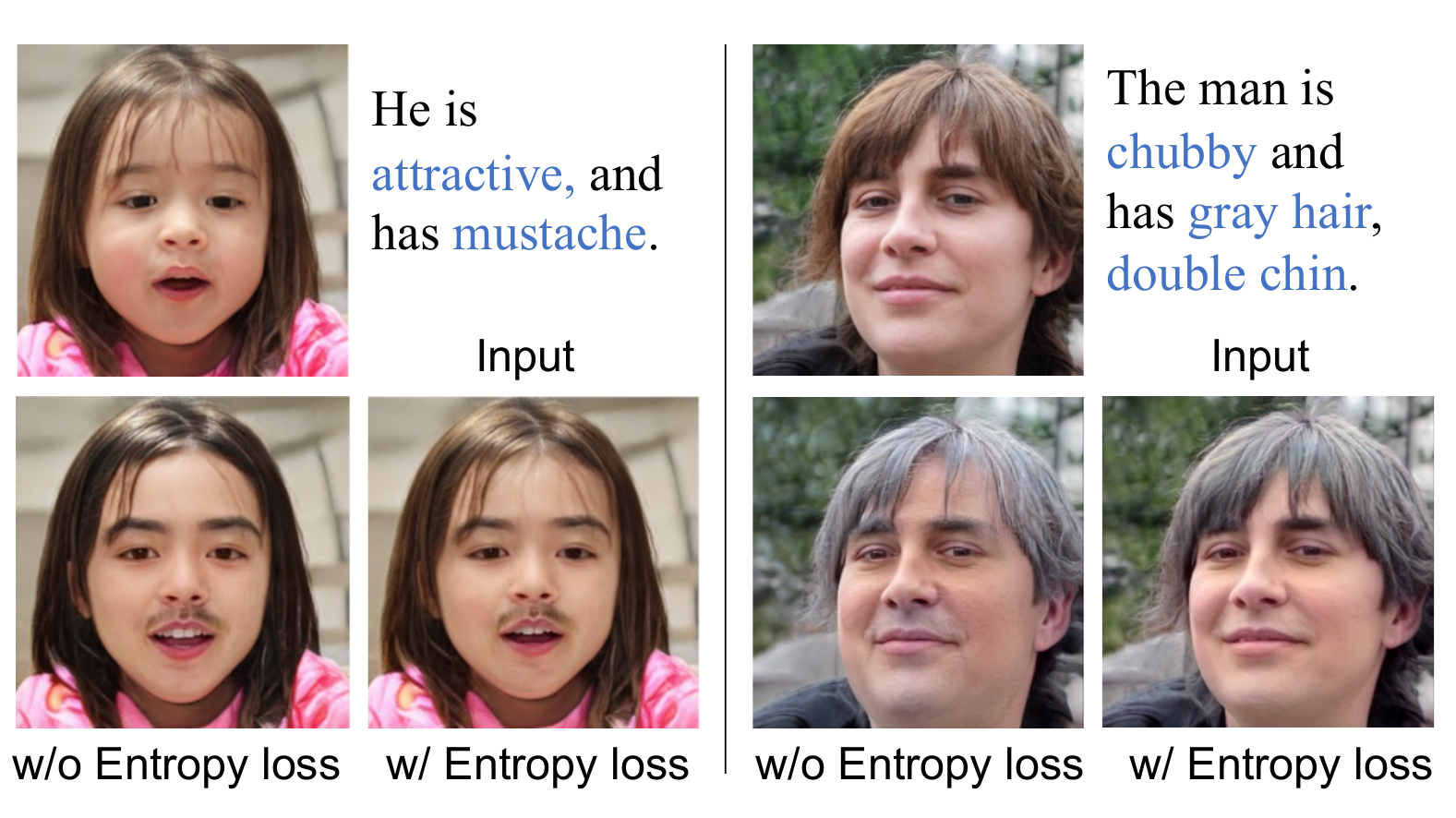}
\end{center}
\vspace{-15pt}
  \caption{\textbf{Effects of our proposed entropy loss.} The model without entropy loss generates images with text-irrelevant edited hair color and identity. Using entropy loss alleviates these issues.}
\vspace{-10pt}
\label{fig:entropy}
\end{figure}

\noindent\textbf{Analysis of Entropy loss.} 
In Figure \ref{fig:entropy}, we present the effects of using the proposed entropy loss during the training phase. To be specific, we input two different textual descriptions to edit the given input image. When we do not adopt the entropy loss, we can see the first edited instances present edited hair color, and the second edited instance has changed the human identity, which are text-irrelevant attributes. On the other hand, the model trained with entropy loss gives better results, since it generates images more similar to the input image in terms of the text-irrelevant attributes, such as hair colors.

\begin{figure}
\begin{center}
\includegraphics[width=0.45\textwidth]{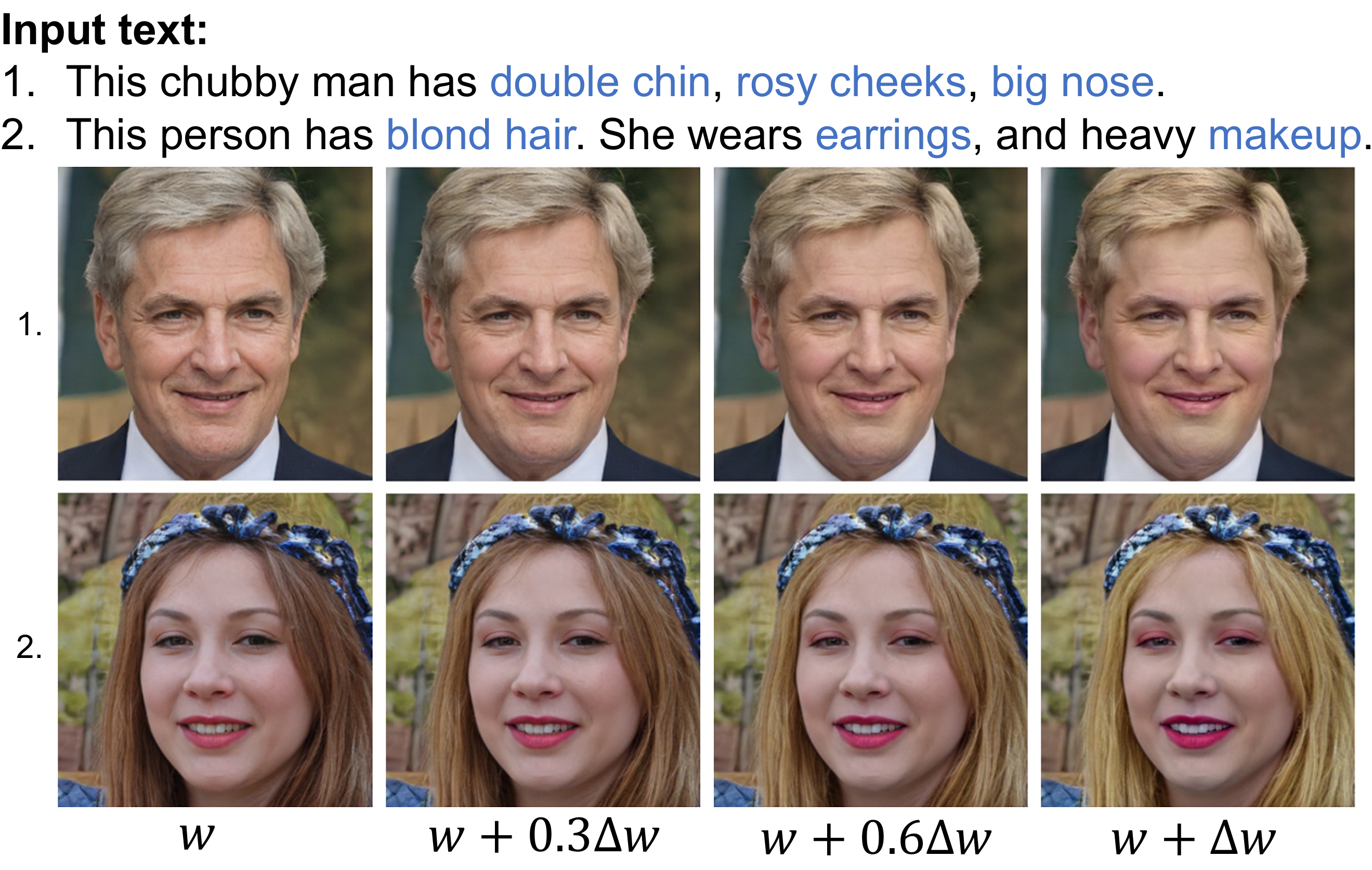}
\end{center}
\vspace{-10pt}
  \caption{\textbf{Results of different weights on $\Delta\w$.} We show the original images and edited images generated from $\w+0.3\Delta\w$, $\w+0.6\Delta\w$ and $\w+\Delta\w$ respectively, which demonstrate editing degrees can be controlled by weights on $\Delta\w$.}
\vspace{-10pt}
\label{fig:offset_weight}
\end{figure}

\begin{figure}
\begin{center}
\includegraphics[width=0.48\textwidth]{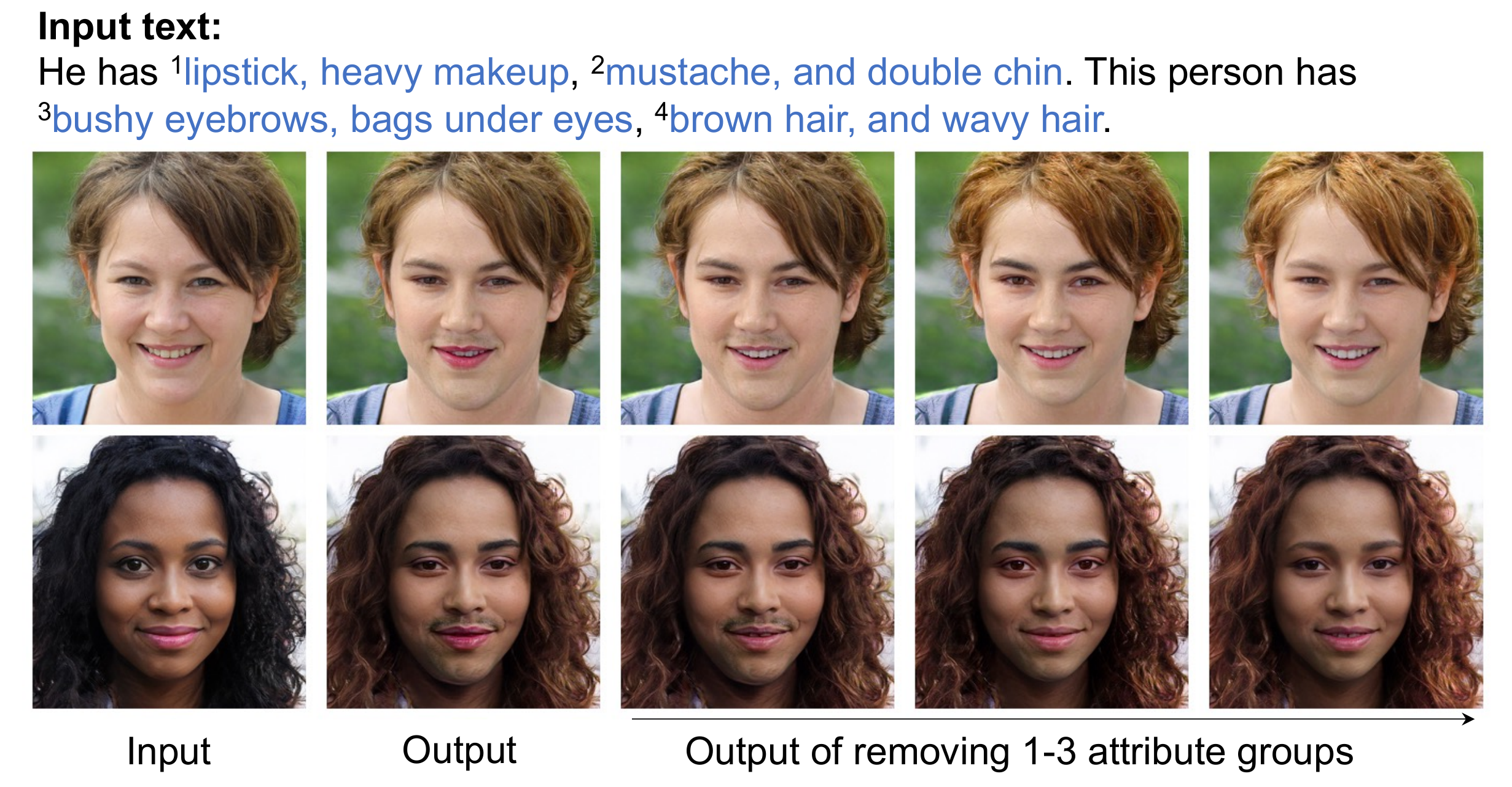}
\end{center}
\vspace{-15pt}
  \caption{\textbf{Results of various attribute numbers.} We input a textual description with 8 attributes, which are categorized into 4 groups. We show ablative results of removing different portions of the input text.}
\vspace{-10pt}
\label{fig:extreme}
\end{figure}

\noindent\textbf{Analysis of different weights on $\Delta\w$.} 
In Figure \ref{fig:offset_weight}, we present the edited images from $\w+0.3\Delta\w$, $\w+0.6\Delta\w$ and $\w+\Delta\w$ respectively. Different weights on $\Delta\w$ give different editing degrees on the manipulated face attributes. It validates the correctness of our model learned editing directions, and produces editing variations on the given input text and images. This increasing editing intensity also shows some potential applications, where users may determine the weights on $\Delta\w$ themselves based on their preferences. 

\noindent\textbf{Ablative results of various attribute numbers.} In Figure \ref{fig:extreme}, we input a long textual description containing 8 different attributes, which are categorized into 4 groups. To validate the efficacy of our face editing method from text with different attribute numbers, we show the ablative edited results based on text having 8 to 2 attributes. We first present the edited images based on the full description, then we gradually remove 1-3 attribute groups and produce the corresponding edited images. Generally, we observe our edited results are natural and matched with the given text, without text-irrelevant change. However, when we change the input attribute number, the editing degrees are different regarding the same attribute. For example, the edited hair color become more \emph{brown} when editing less attribute numbers. This correlation indicates the limitation of our method, which is our method cannot control the editing degree for each attribute individually, since the latent code offsets are produced based on all attributes.

\subsection{Celebrity image manipulation} \label{sec:cele}
\begin{figure}
\begin{center}
\includegraphics[width=0.45\textwidth]{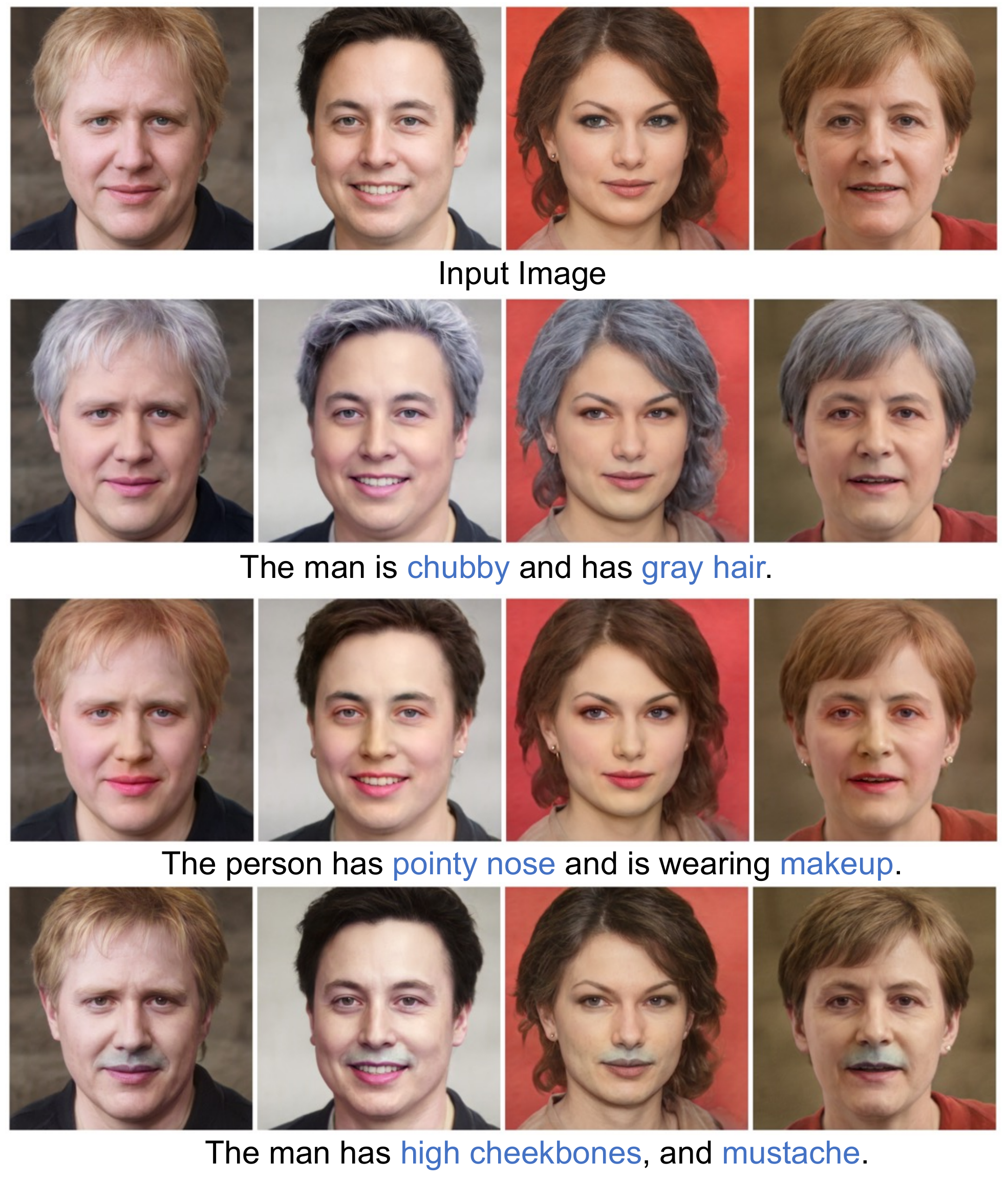}
\end{center}
\vspace{-15pt}
  \caption{\textbf{Manipulation results of celebrity face images.} The latent codes are obtained by GAN inversion \cite{tov2021designing}. In the 2-4 rows, we show the edited images given different text inputs.}
\vspace{-15pt}
\label{fig:real}
\end{figure}

In Figure \ref{fig:real}, we show manipulation results over celebrity face images, where the images are extracted from StyleCLIP \cite{patashnik2021styleclip} paper and we use e4e encoder \cite{tov2021designing} to obtain the corresponding latent codes. We show the edited results based on three different sentences. It can be seen our edited results have various editing intensity for different faces, which fits each image and makes it look natural. Moreover, our manipulated images preserve the identity information and fine-grained details well, indicating the disentanglement property of our model. 
For example, when we attempt to add \emph{mustache} onto female faces in the edited images, other face attributes such as \emph{hair} and \emph{earrings} remain unchanged. However, our model fails to give good results on the gender manipulation. Given the text of \emph{man}, the edited results of the third and fourth columns do not reflect significant changes on the gender. This is because there is a trade off between the identity preservation and gender editing, and our model achieve high ID similarity in the evaluation metrics, which restricts the performance on gender editing.

\section{Conclusion}
In this paper, we have proposed a novel framework, ManiCLIP, that tackles the multi-attribute face editing problem with text input.
To achieve minimal excessive editing, we introduce a decoupling training scheme and add entropy loss to the learning objective. 
As a result, the edited images not only match the given text, but also are natural and have minimal irrelevant attribute change, maximally preserving the original identity. In a series of comprehensive experiments, we show that our proposed method outperforms various baseline models. 

\textbf{Limitation.} Please refer to the last paragraphs of Section \ref{sec:ablation} and \ref{sec:cele} for our limitation discussion.

{\small
\bibliographystyle{ieee_fullname}
\bibliography{egbib}
}

\end{document}